\documentclass[runningheads]{llncs}
\usepackage[T1]{fontenc}
\usepackage{graphicx}
\usepackage{pifont}
\usepackage{slashbox}
\usepackage{multirow}

\usepackage{booktabs,subcaption,amsfonts,dcolumn}
\begin{document}
\title{Incremental Learning \\on Food Instance Segmentation}
%

\author{Huu-Thanh Nguyen\inst{1} \and
Yu Cao\inst{2} \and
Chong-Wah Ngo\inst{2} \and
Wing-Kwong Chan\inst{1}}
%

%
\institute{Department of Computer Science, City University of Hong Kong, Hong Kong
\email{tnguyenhu2-c@my.cityu.edu.hk,wkchan@cityu.edu.hk}\\
\and
School of Computing and Information Systems, Singapore Management University, Singapore
\email{yu.cao.2022@msc.smu.edu.sg,cwngo@smu.edu.sg}}

\maketitle              
\begin{abstract}
Food instance segmentation is essential to estimate the serving size of dishes in a food image. The recent cutting-edge techniques for instance segmentation are deep learning networks with impressive segmentation quality and fast computation. Nonetheless, they are hungry for data and expensive for annotation. This paper proposes an incremental learning framework to optimize the model performance given a limited data labelling budget. The power of the framework is a novel difficulty assessment model, which forecasts how challenging an unlabelled sample is to the latest trained instance segmentation model. The data collection procedure is divided into several stages, each in which a new sample package is collected. The framework allocates the labelling budget to the most difficult samples. The unlabelled samples that meet a certain qualification from the assessment model are used to generate pseudo-labels. Eventually, the manual labels and pseudo-labels are sent to the training data to improve the instance segmentation model. On four large-scale food datasets, our proposed framework outperforms current incremental learning benchmarks and achieves competitive performance with the model trained on fully annotated samples.


\keywords{Food Computing \and Incremental learning  \and Instance segmentation \and Semi-supervisor learning \and Membership inference.}
\end{abstract}
\section{Introduction}
Instance segmentation is a fundamental task in computer vision with several applications, including food portion size estimation \cite{2017grab,2021alan,sibnet}, text localization \cite{2019Wenhai,2021mountain}, vehicle surveillance \cite{2017Bai,2019Neven}. With the development of deep learning, several neural networks like Mask R-CNN \cite{2017maskrcnn}, CenterMask \cite{2020centermask}, Watershed \cite{2017Bai}, Terrace \cite{2021alan}, have been developed to address the problem. Nonetheless, these techniques prioritize segmentation quality and computation efficiency over the annotation effort. This paper aims to fill this gap with the proposal of an incremental learning framework that maximizes the competence of the labour force by automatically selecting the most difficult samples for annotation and high-quality pseudo-labels from the remaining samples for model improvement.
 
\begin{figure}[t]
\centering
\includegraphics[width=0.66\textwidth]{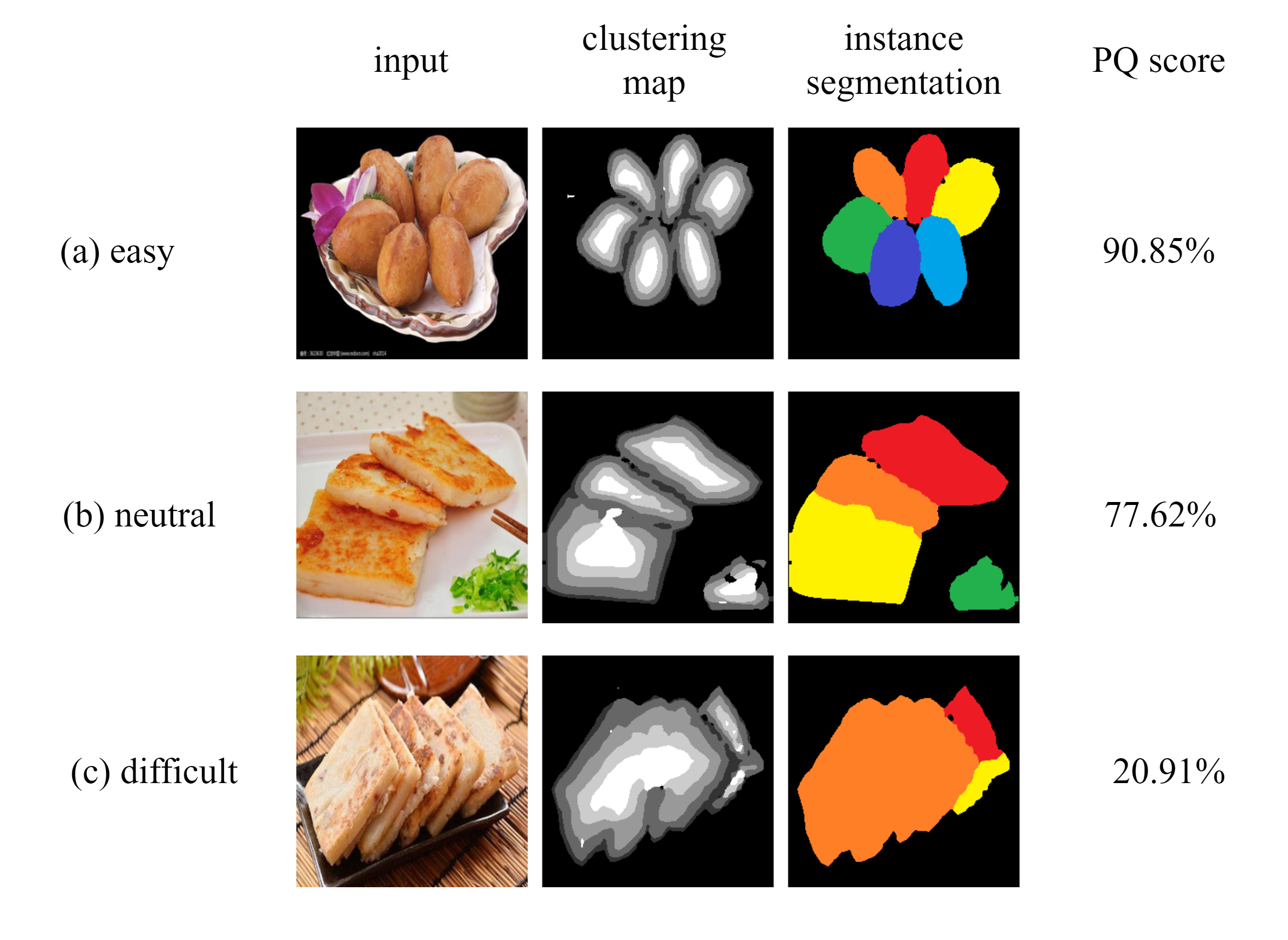}
\caption{Inferences of the latest instance segmentation model are made on new collected samples. For each sample, the model outputs a clustering and an instance map. The PQ \cite{2018panoptic} score is calculated from the generated and the actual instance maps.} \label{fig:intro}
\end{figure}

Over the past few years, food computing has received great attention with several research papers on food recognition \cite{2017food}, detection \cite{2019deng} and segmentation \cite{2021alan}. Among these tasks, food instance segmentation is essential for portion size estimation \cite{2021pose}. However, preparing ground-truth instance segmentation masks for model training is time-consuming. To address the annotation problem, most existing strategies primarily reduce the amount of supervised information, ranging from image-level \cite{2018WeaklySI}, instance-level \cite{2018issam}, bounding box \cite{2019deng}, and polygon \cite{2021alan,sibnet}, introducing a trade-off between segmentation performance and annotation effort. Instead of an instance mask, the image-level and instance-level may either generate a peak response map or a blob that provides the instance locations in the input image. Bounding box information is better for estimating instance size but fails to predict the instance shape. Polygon supervision exhibits the most appropriate technique that annotates critical corner points on the instances to secure segmentation quality. However, for a complex appearance, such as in the food domain, the large number of critical points leads to expensive polygon annotation. For such a situation, semi-supervision \cite{budget} is a potential solution where some random samples are annotated, and the remaining ones are prepared with model-generated instance masks. A risky aspect of this approach is that the quality of the generated masks is not guaranteed for model improvement.

This paper presents an incremental learning framework for instance segmentation, with the novelty being the proposal of an assessment module for automatically scoring the difficulty of newly collected samples. In practice, data samples are collected gradually in package units for research and commercial purposes. Our framework annotates and trains an initial instance segmentation model on the first package. In parallel, we use this package to train a difficulty assessment model that evaluates quantitatively how much each new sample is challenging to the latest instance segmentation model. Usually, a new package comprises easy, neutral, and hard samples, as illustrated in Fig. \ref{fig:intro}. The ready-to-use assessment model is employed to forecast their difficulty levels. With inspiration from hard-mining \cite{hard_mining} and self-paced learning \cite{selfpace}, we then direct hard samples to the labour force for polygon annotation and the easy samples to generate pseudo-labels. The neutral and easy samples will then be preserved in pooling data for consideration in the next stage. The manual labels and high-quality pseudo-labels are used to fine-tune the instance segmentation model. We repeat this procedure when either a new sample package is collected or the labour force is available for the annotation work.

\section{Related Work}
\textbf{Instance segmentation} Current mainstream deep learning techniques for instance segmentation can be classified into two major groups: proposal-based \cite{2017maskrcnn,2020centermask} and clustering-based \cite{2021alan,2017Bai}. Proposal-based techniques, such as Mask R-CNN \cite{2017maskrcnn} and Center Mask \cite{2020centermask}, generate a set of candidate bounding boxes for instance detection, and following each box is a prediction for the corresponding instance mask. Clustering-based generates one or many instance feature maps, such as multiple terrace layers \cite{2021alan}, watershed energy and centre-direction maps \cite{2017Bai}, that can construct an instance segmentation map. This paper exploits Terrace \cite{2021alan}, a clustering-based instance segmentation technique which exhibits impressive performance in the food domain, for our incremental learning framework. The use of the Terrace technique is two-fold. First, given a new sample, the output is a terrace probability distribution, representing how certain the model is about its prediction. This paper investigates the generated probability distribution to forecast the difficulty level of a new image sample. Second, we employ Terrace \cite{2021alan} as the core instance segmentation model to develop and evaluate the proposed incremental learning framework.

\textbf{Semi-supervision instance segmentation} To improve the instance segmentation performance with a limited human resource budget for image annotation, \cite{budget} proposes a semi-supervised learning mechanism. There are two instance segmentation models proposed to work in sequence: an annotation model and a production model. When a new package of samples is collected, a small number of samples will be randomly selected and manually labelled. The annotation model is trained on samples with manual labels and then used to generate pseudo-labels for the remaining unlabelled samples. The production model is trained on both manual and pseudo-labels. \cite{budget} argues that the extra knowledge from pseudo-labels improves the segmentation performance at no additional cost. However, the generated pseudo-labels are not properly qualified. While a good pseudo-label may provide more context to support the model in distinguishing instance versus background pixels to be aware of instance boundaries, a bad pseudo-label may give the wrong information and confuse the production model. In our framework, we propose a difficulty assessment technique to forecast the quality of generated pseudo-labels. Then, only samples with good pseudo-labels are used to improve the production model, while the remaining samples can be either skipped or annotated depending on labour force availability.

\textbf{Membership inference attack} Given a deep learning model with open access, \cite{attack} proposes a technique to predict whether a query sample is inside or outside the training set. An incremental learning framework may employ this technique to filter out the used samples and spend the annotation budget on new samples. This paper proposes a more advanced method, which may forecast the panoptic quality (PQ) \cite{2018panoptic} score of every new sample (without ground-truth instance segmentation mask), reflecting how much the sample is challenging to the latest instance segmentation model.

\section{Incremental Learning Framework}
\begin{figure}[!t]
\includegraphics[width=\textwidth]{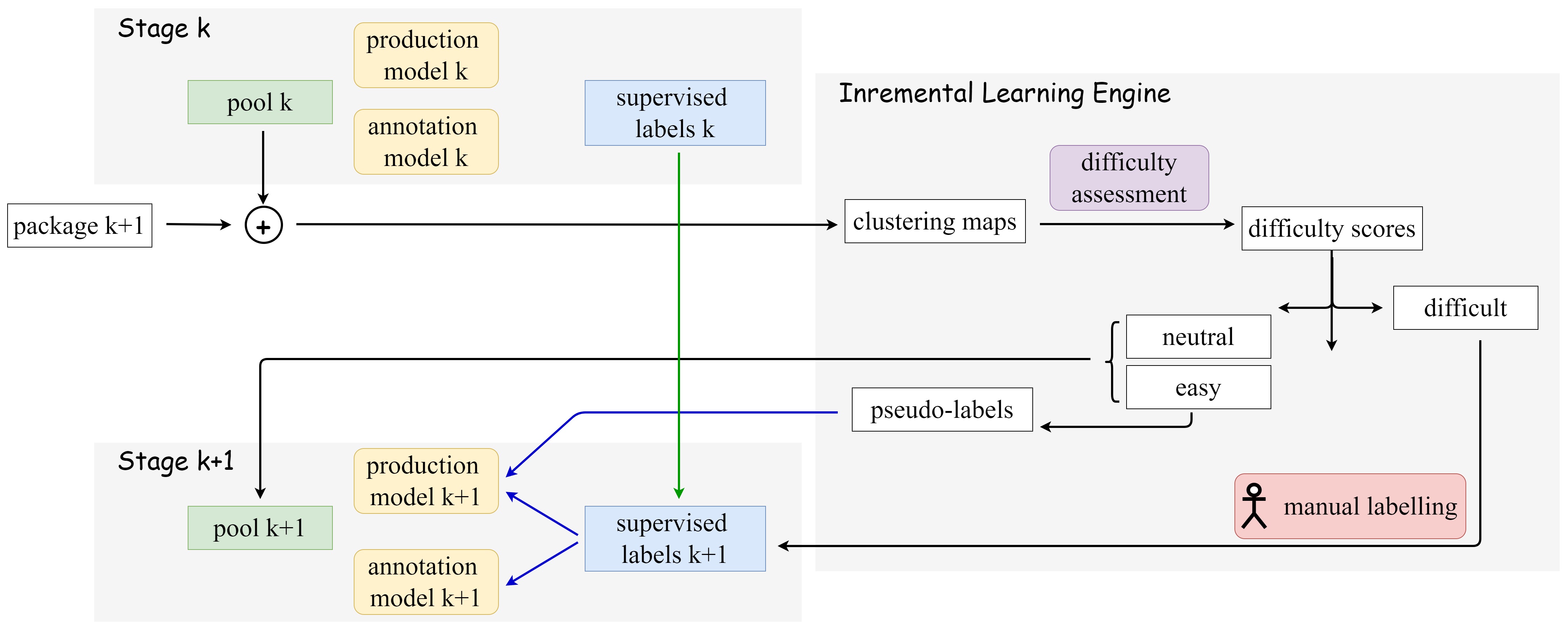}
\caption{Incremental learning framework continuously updates newly collected data and improves the instance segmentation model.} \label{fig:arch}
\end{figure}

Fig. \ref{fig:arch} illustrates the proposed incremental learning on instance segmentation framework, including an incremental learning engine interconnecting the states of collected samples, supervised labels, and instance segmentation models between successive stages. In the scope of this paper, we employ the clustering-based technique Terrace \cite{2021alan} for instance segmentation. At a specific stage, our system stores unlabelled samples in a data pool, a set of supervised samples, an annotation model, and a production model. The incremental learning engine receives unlabelled samples, forecasts their challenging scores, and navigates them to the manual labelling module, pseudo-labelling module, or holds them in the data pool for future use. 

\subsection{Data Pool, Labels and Instance Segmentation Model}
In both academia and industry, the data collection process is either automatic by crawler tools or manual by humans. As the required data for machine learning is usually large-scale, it takes a timely basis, such as daily, weekly, or monthly, for data collection and annotation. This paper defines such a timely basis as a ``stage'' and presents the incremental learning framework between consecutive stages. At stage $k$, the following states are recorded in the learning system::
\begin{itemize}
\item[\ding{32}] \textbf{Data pool} is to store the set of samples where the annotation for instance segmentation masks has not been prepared yet. As the task is time-consuming, there is always a long queue of unlabelled samples. The pool holds the samples unlabelled in the previous stage for use in the current stage.

\item[\ding{32}] \textbf{Supervised labels} is to hold the set of pairs (sample, instance segmentation label). In the proposed framework, we annotate most of the samples collected in the first stage. In the subsequent stages, we eliminate the labelling cost by only selecting a small percentage of samples to annotate and navigating them to the supervised labels. In other words, the expense of the labour force is only expensive for the first stage and can be flexibly adjusted in the subsequent stages.

\item[\ding{32}] \textbf{Annotation model}, inspired from \cite{budget}, is an instance segmentation model trained on supervised samples. At the beginning of each stage, the annotation model makes inferences on all pooling samples to generate clustering maps and pseudo-labels. At the end of the stage, the model is improved when the set of supervised samples is updated.

\item [\ding{32}] \textbf{Production model} shares the same architecture as the annotation model, but it is trained on supervised labels and pseudo-labels. As pseudo-labels are generated from a large-scale data pool, they contain a diverse food context. This diversity supports the production model to understand the data population better. Unlike \cite{budget}, we do not include all the pseudo-labels to improve the production model. Instead, we pick high-quality pseudo-labels, which are automatically recommended by the difficulty assessment module.
\end{itemize}

When a new package $k+1$ is collected, we merge it with the current pooling data $k$ and feed it into the incremental learning engine.


\subsection{Incremental Learning Engine}
The novelty of this paper is the proposal of an incremental learning engine that forecasts the difficulty level of collected samples and distributes them to manual annotation, pseudo-label generation, and data pool. In particular, the samples collected in package $k$ and current samples in the data pool $k$ are merged. Here, each sample is subjected to the latest annotation model, generating a set of clustering maps. Each clustering map compromises a pre-defined number of terrace layers. The probability distribution on terrace layers reveals how the annotation model is confident with its prediction of the sample. Therefore, we explore the outputted terrace layers to train a difficulty assessment model as follows:

\begin{figure}[!t]
\centering
\includegraphics[width=0.8\textwidth]{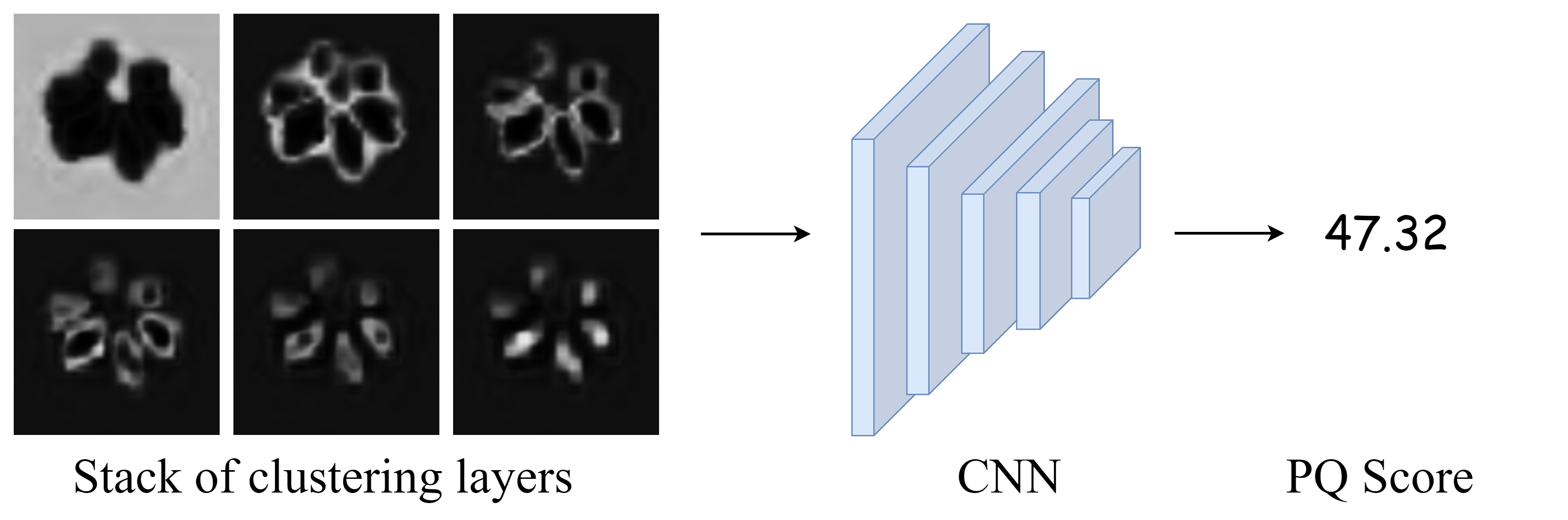}
\caption{Architecture of the difficulty assessment model.} \label{fig:arch_pq}
\end{figure}

\begin{figure}[!t]
\includegraphics[width=\textwidth]{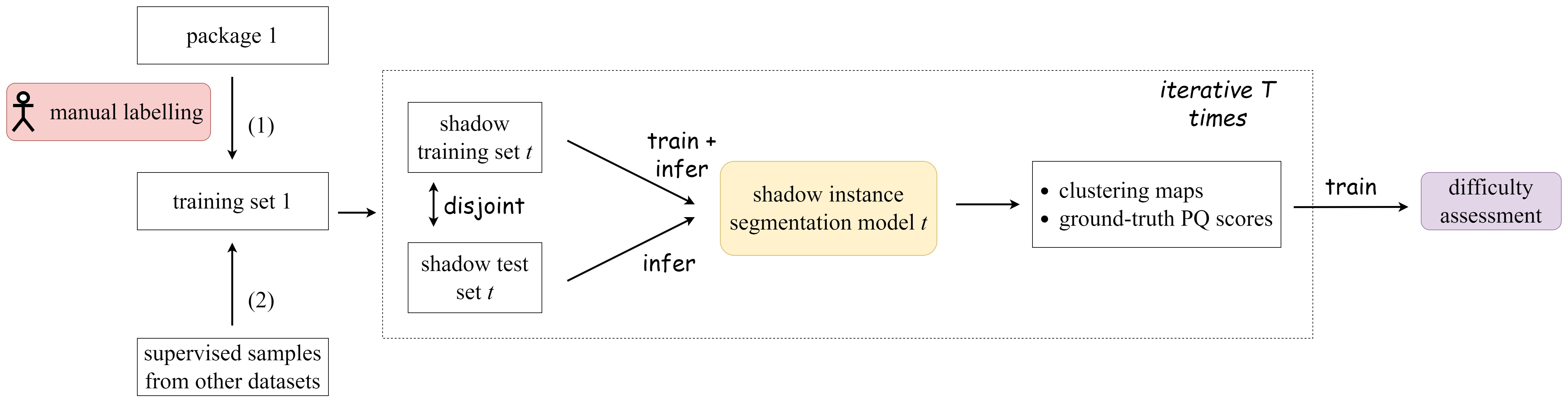}
\caption{Pipeline to generate training clustering maps and ground-truth PQ scores for the difficulty assessment model.} \label{fig:diff}
\end{figure}

\begin{itemize}
\item [\ding{32}] \textbf{Difficulty assessment model} The proposed difficulty assessment model, as shown in Fig. \ref{fig:arch_pq},  is a convolutional neural network composed of convolutional and fully connected layers. This paper employs ResNet-50 \cite{2016resnet}, pre-trained on ImageNet \cite{imagenet}, in experiments. The model's input is a stack of clustering layers representing the predicted terrace probability distributions by the latest annotation model. The model's output is a regression value within the range $[0, 1]$, describing how the model forecasts the PQ score of a new sample made on the annotation model. A low score means a difficult sample, and a high score means an easy sample. The loss function for difficulty assessment model training is formulated as follows:
\begin{equation}
L = \frac{\sum^{N}_{n=1}(s_n - \hat{s}_n)^2}{N}
\end{equation} 
where $s_n$ and $\hat{s}_n$ are ground-truth and predicted PQ scores for $n^{th}$ sample, and $N$ is the total number of training samples.

\item [\ding{32}] \textbf{Generating training samples} Inspired by the membership attack framework \cite{attack}, we generate training samples for the difficulty assessment model via shadow models, as illustrated in Fig. \ref{fig:diff}. First, a sample package is collected and manually annotated. In transfer learning, an alternative solution is to use annotated samples from the ready-to-use datasets. Second, the labelled samples are distributed into shadow pairs of training and testing sets. Third, each shadow training set is used to train a shadow Terrace model for instance segmentation, which shares the same hyper-parameters (e.g., number of terrace layers) as the production model. It is noted that all these samples, both for training and testing, are prepared with ground-truth instance segmentation masks. Therefore, the generated instance maps from a shadow model can be measured with PQ scores. A predicted clustering map and the corresponding PQ score form a pair (clustering map, PQ score). We repeat the second and the third steps to build up large-scale (clustering map, PQ score) samples to train and test the difficulty assessment model. We iterated the process 80 times in the experiment, resulting in over 35,000 pairs of (clustering map, PQ score). We select roughly 9,100 pairs in a uniform distribution for the PQ score for the assessment model training and evaluation.
\end{itemize}

To this end, the proposed incremental learning framework automatically forecasts the challenging level of a new sample by feeding it into the latest annotation model and the difficulty assessment model. Depending on the budget available, a percentage of the most challenging samples is selected for manual labelling. The remaining samples are navigated into neutral and easy sets, where the easy samples are required to meet a certain threshold of difficulty score. The generated instance maps of easy samples on the latest annotation model are used as pseudo-labels. For the next production model, low-weighted pseudo-labels and high-weighted supervised labels are sent to the training set. In particular, we weigh the samples in the first package, the hard and the easy samples in a ratio of 2:4:1 to better mine the challenging cases. For the annotation model, only supervised labels are used. Finally, neutral and easy samples are put into the data pool for use in the next stage.

\section{Experiments}
\subsection{Experimental Setup}
\subsubsection{Dataset} Four food datasets, including Dimsum, Sushi, Cookie, and UECFoodPixComp (UEC) \cite{2021alan,uecfoodpixcomplete}, are employed to evaluate the incremental learning performance. While Sushi is only used as a pre-trained dataset to evaluate various sampling strategies for transfer learning on the Dimsum dataset, each of the remaining datasets is split into six packages to evaluate the incremental learning framework. It is noted that the number of food instances is different among image samples. To be fair for the instance segmentation labelling budget, which is instance-based, we organize the packages with a similar number of food instances: 1,575. The number of samples for evaluation in Dimsum, Cookie, and UEC is 768, 1152, and 1000, respectively. 

\subsubsection{Evaluation Metric} Regarding instance segmentation performance, we employ Panoptic Quality (PQ) \cite{2018panoptic} to evaluate the performance of the production model. For the difficulty assessment model, which outputs a PQ-based difficulty score, we measure the absolute error between the predicted and ground-truth PQ scores. To justify the efficiency of the proposed incremental framework, we evaluate the instance segmentation performance across various settings of manual labelling efforts, including 5\%, 10\%, 20\%, 30\% and 100\% annotation time for each data package.

\begin{table}[!t]
\centering
\caption{The instance segmentation performance (\% PQ) with 5\% and 10\% training samples labelled by various data sampling strategies on the Dimsum dataset. Transfer learning is made with a model pre-trained on Sushi dataset.}\label{tab:tab1}
\begin{tabular}{|c|cc|ccc|}
\hline
\hline
Sampling strategy       & \multicolumn{2}{c|}{Trivial learning} & \multicolumn{3}{c|}{Transfer learning}                          \\ \hline
      & \multicolumn{1}{c|}{5\%}     & 10\%   & \multicolumn{1}{c|}{0\%}   & \multicolumn{1}{c|}{5\%}   & 10\%  \\ \hline
easy   & \multicolumn{1}{c|}{45.66}   & 54.79  & \multicolumn{1}{c|}{76.33} & \multicolumn{1}{c|}{76.88} & 77.54 \\ \hline
random & \multicolumn{1}{c|}{58.77}   & 62.75  & \multicolumn{1}{c|}{76.33}      & \multicolumn{1}{c|}{79.52} & 80.74 \\ \hline
hard   & \multicolumn{1}{c|}{\textbf{62.99}}   & \textbf{65.96}  & \multicolumn{1}{c|}{76.33}      & \multicolumn{1}{c|}{\textbf{80.61}} &\textbf{82.84} \\ \hline \hline
\end{tabular}
\end{table}
\subsection{Experimental Results}
\subsubsection{Data sampling for instance segmentation} First, we investigate easy, random, and hard sampling strategies for food instance segmentation. Table \ref{tab:tab1} lists their performances on the Dimsum dataset. The hard-mining strategy consistently outperforms the easy and random sampling in both trivial and transfer learning with a gap of over 1-4\% of PQ score when training on only 5\% training samples and 1-3\% of PQ score on 10\% training samples. When the annotation budget is limited, it is better to prioritize the labour resources for the most challenging samples.

\begin{figure}[!t]
\centering
\includegraphics[width=\textwidth]{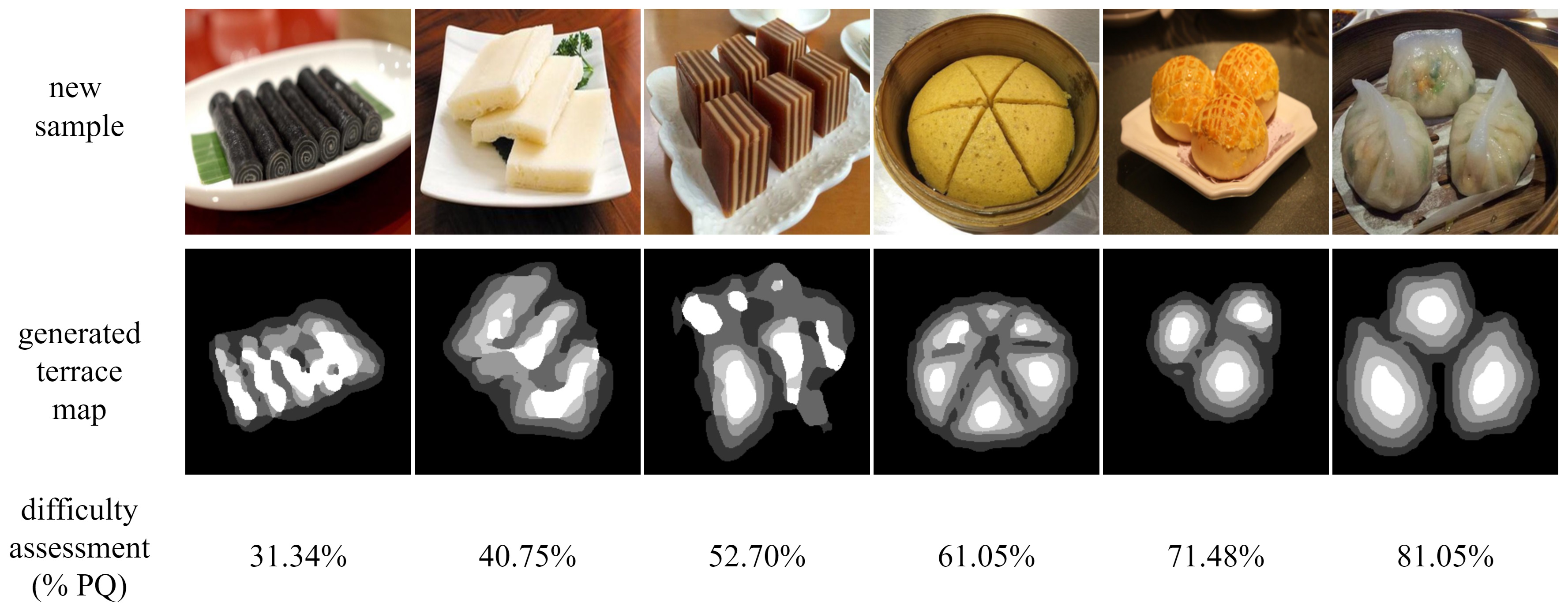}
\caption{The forecast of difficulty level on unlabelled samples. A lower predicted PQ score means a more difficult sample.} \label{result:assessment}
\end{figure}
\subsubsection{Difficult assessment model} We evaluate the assessment model by the mean absolute error on the validation set presented in Section 3.2. The recorded error is relatively low, at 7.8\%. Therefore, the model is promising in forecasting the difficulty score for an unlabeled sample. Fig. \ref{result:assessment} visualizes some food images and the predicted PQ scores. On the first three samples with the challenges due to the obscure boundary, stacking, and occlusion problems, the assessment model gives lower PQ scores, meaning challenges to the current instance segmentation model. Meanwhile, higher scores, considered easy samples, are given to the last three samples, where the boundaries among instances are relatively clear.

\begin{table}[!t]
\centering
\caption{The trade-off between the quality and quantity of pseudo-labels to incremental learning .}\label{tab:threshold}
\begin{tabular}{|c|c|c|c|c|c|}
\hline
\hline
threshold for easy samples & 0     & 20    & 40    & 60             & 80    \\ \hline
performance                & 82.52\% & 82.85\% & 83.06\% & \textbf{83.34\%} & 83.19\% \\ \hline
\hline
\end{tabular}
\end{table}

\subsubsection{Pseudo-labels} Next, we examine how pseudo-labels contribute to incremental learning. Table \ref{tab:threshold} lists the instance segmentation quality on the Dimsum dataset given different thresholds to accept pseudo-labels. There is a trade-off between the pseudo-label quality and the number of easy samples explored. On the one hand, a low-quality threshold (e.g., at 0\% or 20\% PQ scores), proposed by \cite{budget}, lets the model explore the context from a large number of unlabelled samples. On the other hand, a high-quality threshold (e.g., at 80\% PQ score) guarantees the model learns from more precise pseudo-labels. The medium PQ threshold, at 60\%, is recorded as a balancing point to retrieve pseudo-labels for incremental learning.

\subsubsection{Incremental Learning strategies} We compare the proposed incremental learning strategies, namely PQ-based$^\ast$, against various benchmarks of random and hard sampling techniques on the Dimsum dataset. Random sampling shuffles the newly collected samples and navigates 10\% of them to the human resource for annotation. The advanced method, random$^\ast$ \cite{budget}, employs the latest annotation model to generate pseudo-labels for the remaining 90\% samples. Hard sampling techniques forecast the difficulty scores for every sample in the new package and give a higher annotation priority to the more difficult samples. Membership-based method \cite{attack} evaluates whether a new sample is similar to one of the existing members in the current training data. While the output label of this method is either yes or no to the question, the output confidence score can be used as an indicator to assess how challenging the samples are. However, as membership inference is a binary classification problem, the confidence score distribution is skewed to 0\% for non-membership and 100\% for membership. Our techniques, PQ-based and PQ-based $^\ast$, generate difficulty scores in a more balanced distribution. The PQ-based $^\ast$ method adds pseudo-labels to training samples. Different from random$^\ast$ \cite{budget}, we can exclude low-quality pseudo samples thanks to the proposed assessment model. 

Table \ref{tab:comparison} lists the performance of these incremental learning strategies on the Dimsum dataset, where the annotation budget is used for only 10\% of the new sample package. Overall, PQ-based$^\ast$ consistently outperform the remaining strategies across six stages. The 10\% package in hard sampling techniques, membership-based and PQ-based, contributes to around 1.5\% improvement in panoptic quality compared to random sampling. PQ-based demonstrates a better assessment approach than membership-based, especially at the early stages. PQ-based$^\ast$, aggregating the advantages of hard sampling and good pseudo-labels, make improvements of 1.5\%-2\% compared to the benchmark random$^\ast$ \cite{budget}.

\begin{table}[!t]
\centering
\caption{The instance segmentation performances (\% PQ) of various incremental learning techniques on the Dimsum dataset. For a new data package, only 10\% of food instances are selected to prepare ground-truth instance segmentation masks.}\label{tab:comparison}
\begin{tabular}{|c|c|cc|ccc|}
\hline
\hline
\multirow{2}{*}{package} &  full annotation       & \multicolumn{2}{c|}{random sampling}                  & \multicolumn{3}{c|}{hard sampling}                                                  \\ \cline{2-7} 
                         &         & \multicolumn{1}{c|}{random}  & random$^\ast$ \cite{budget}& \multicolumn{1}{c|}{membership \cite{attack}} & \multicolumn{1}{c|}{PQ-based}      & PQ-based$^\ast$ \\ \hline
1                        & 74.60 & \multicolumn{1}{c|}{}        &                        & \multicolumn{1}{c|}{}           & \multicolumn{1}{c|}{}        &                    \\ \hline
2                        &         & \multicolumn{1}{c|}{76.23} & 76.73                & \multicolumn{1}{c|}{76.57}    & \multicolumn{1}{c|}{77.08} & \textbf{78.32}   \\ \hline
3                        &         & \multicolumn{1}{c|}{77.30} & 78.81                & \multicolumn{1}{c|}{78.42}    & \multicolumn{1}{c|}{79.49} & \textbf{80.98}   \\ \hline
4                        &         & \multicolumn{1}{c|}{79.39} & 79.87                & \multicolumn{1}{c|}{80.57}    & \multicolumn{1}{c|}{81.24} & \textbf{81.99}   \\ \hline
5                        &         & \multicolumn{1}{c|}{80.27} & 80.47                & \multicolumn{1}{c|}{80.97}    & \multicolumn{1}{c|}{81.90} & \textbf{82.74}   \\ \hline
6                        &         & \multicolumn{1}{c|}{80.75} & 81.95                & \multicolumn{1}{c|}{82.11}    & \multicolumn{1}{c|}{82.39} & \textbf{83.34}   \\ \hline
\hline
\end{tabular}
\end{table}

\begin{figure}[!t]
    \centering
    \begin{subfigure}[b]{0.32\textwidth}
        \includegraphics[width=\textwidth]{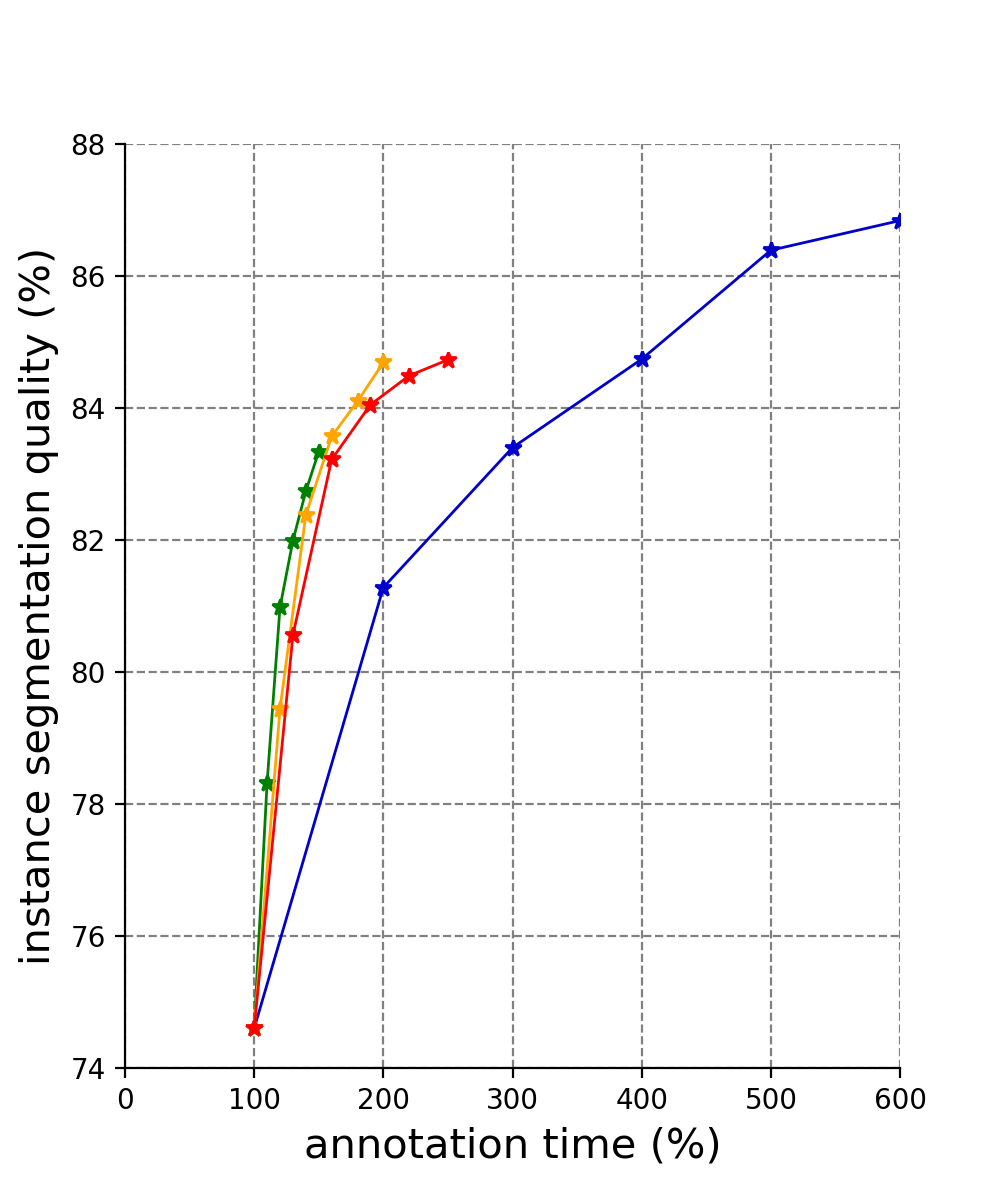}
        \caption{Dimsum}
    \end{subfigure}
    \begin{subfigure}[b]{0.32\textwidth}
        \includegraphics[width=\textwidth]{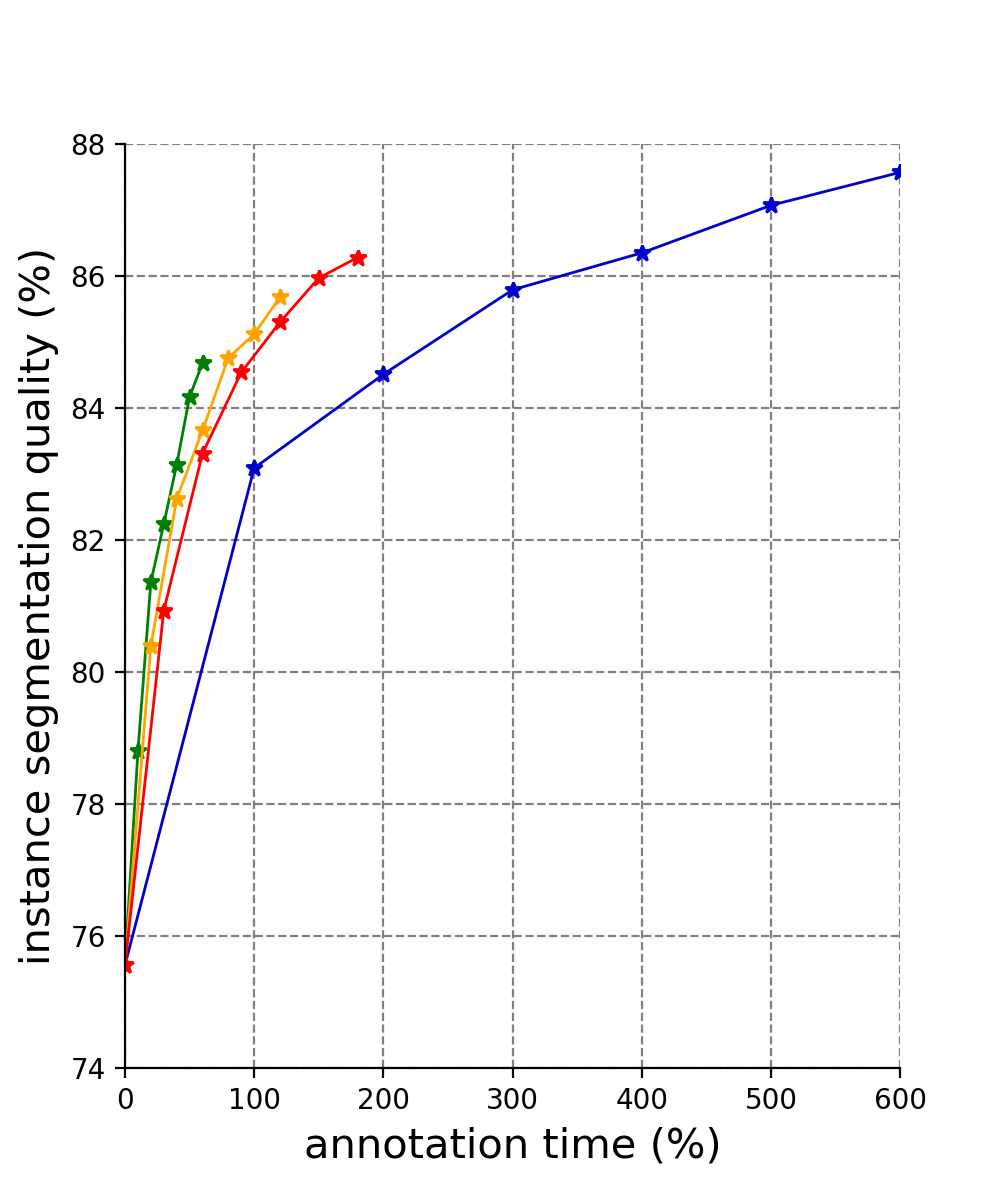}
        \caption{Cookie}
    \end{subfigure}
    \begin{subfigure}[b]{0.32\textwidth}
        \includegraphics[width=\textwidth]{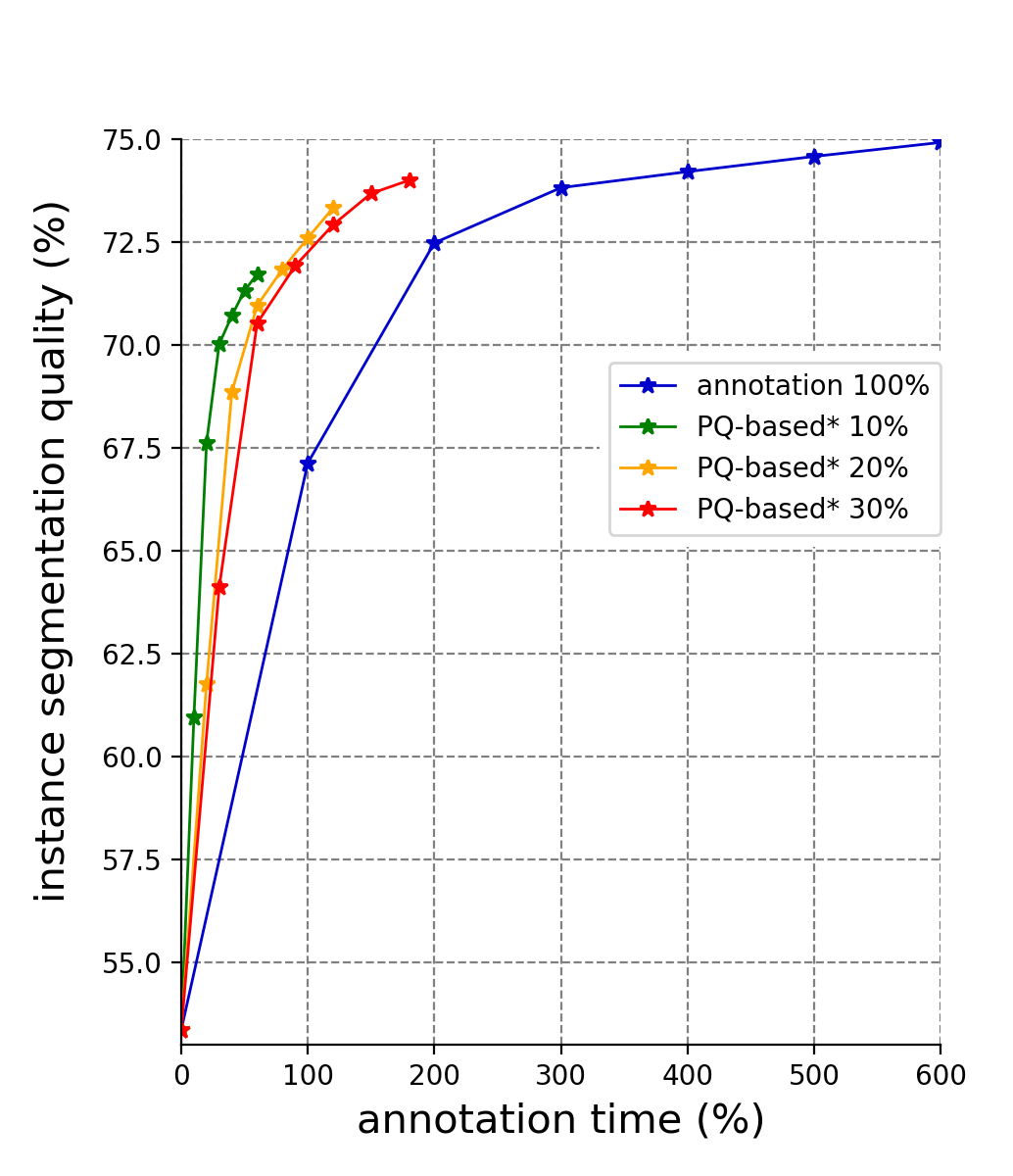}
        \caption{UEC}
    \end{subfigure}
    \caption[01]{Incremental learning: labelling effort and instance segmentation performance.}
    \label{fig:transfer}
\end{figure}

\subsubsection{Annotation effort and model performance} Last, we compare model performances given by the proposed method, PQ-based$^\ast$, on a little annotation effort and the traditional approach training on manual labels for all collected samples. Fig. \ref{fig:transfer}a illustrates the panoptic quality over annotation time on the Dimsum dataset. Given the pre-trained model PQ-based$^\ast$ 10\% on Dimsum, we perform transfer learning on Cookie and UEC datasets. On the Dimsum dataset, at the last package, with only one-fourth of the annotation effort, PQ-based$^\ast$ 10\% is competitive with the fully annotated approach. The approaches PQ-based$^\ast$ 20\% and PQ-based$^\ast$ 30\%, with one-third and one-half of the annotation time, can achieve equivalent performance to the model with 100\% data annotated. Using the pre-trained model of PQ-based$^\ast$ 10\% on Dimsum, the transfer learning models of PQ-based$^\ast$ 30\% on Cookie and UEC show a minor gap of around 1\% PQ less than the model with 100\% sample annotated. 

\section{Conclusion}
We have presented an incremental learning framework for food instance segmentation given a fixed budget for annotation. The framework's power comes from the proposed assessment model, which forecasts the difficulty scores for unlabelled samples. The score enables the framework to select hard samples for manual labelling and high-confidence samples for pseudo labels. The experimental results justify the efficiency of the assessment model in scoring food images regarding how challenging they are to instance segmentation. The proposed framework outperforms current incremental learning benchmarks on the Dimsum dataset with the same amount of annotation effort. The framework exhibits competitive performance with fully annotated models on the Dimsum dataset and the transfer learning on the Cookie and UEC datasets, with a much shorter labelling time. The proposed incremental learning strategy is a promising solution to train and deploy a food instance segmentation model in practice.



%
%
%
\bibliographystyle{unsrt}
\bibliography{reference}
\end{document}